\documentclass[a4paper,12pt]{article}

\usepackage[utf8]{inputenc} 
\usepackage[T1]{fontenc}    
\usepackage{hyperref}       
\usepackage{url}            
\usepackage{booktabs}       
\usepackage{amsfonts}       
\usepackage{nicefrac}       
\usepackage{microtype}      

\usepackage{color}
\usepackage{natbib}
\usepackage{graphicx}
\usepackage{amsmath}
\usepackage{multirow}

\newcommand{\be}{\begin{equation}}
\newcommand{\ee}{\end{equation}}
\newcommand{\bi}{\begin{itemize}}
\newcommand{\ei}{\end{itemize}}
\newcommand{\bea}{\begin{eqnarray}}
\newcommand{\eea}{\end{eqnarray}}

\newcommand{\bfmu}{\boldsymbol{\mu}}

\newcommand{\bfe}{\mathbf{e}}

\newcommand{\bfs}{\mathbf{s}}

\newcommand{\bfu}{\mathbf{u}}

\newcommand{\bfx}{\mathbf{x}}
\newcommand{\bfy}{\mathbf{y}}
\newcommand{\bfz}{\mathbf{z}}

\newcommand{\bfzero}{\mathbf{0}}

\newcommand{\bbE}{\mathbb{E}}

\newcommand{\cut}[1]{}

\newcommand{\landh}{L\&H}
\newcommand{\matlab}{\textsc{Matlab}}
\newcommand{\tW}{\tilde{W}}
\newcommand{\bfyxi}{\bfy_{\xi}}

\title{
\makebox[0pt][l]{\raisebox{1in}[0pt][0pt]{
\parbox{6in}{\small Final m/s version of paper accepted for
  publication in  \emph{Neural Computation}.}}}  
Fusing Foveal Fixations Using Linear Retinal Transformations
and Bayesian Experimental Design}

\author{Christopher K. I. Williams \\
  School of Informatics \\
  University of Edinburgh, UK \\
\texttt{c.k.i.williams@ed.ac.uk} }

\begin{document}

\maketitle

\begin{abstract}
Humans (and many vertebrates) face the problem of fusing together
multiple fixations of a scene in order to obtain a representation of
the whole, where each fixation uses a high-resolution fovea and
decreasing resolution in the periphery.
In this paper we explicitly represent the retinal transformation of a fixation
as a linear downsampling of a high-resolution latent image of the
scene, exploiting the known geometry.  This linear transformation
allows us to carry out exact inference for the latent variables in
factor analysis (FA) and mixtures of FA models of the scene. Further, this
allows us to formulate and solve the choice of ``where to look next''
as a Bayesian experimental design problem using the Expected
Information Gain criterion. Experiments on the Frey
faces and MNIST datasets demonstrate the effectiveness of our models.
\end{abstract}


\section{Introduction}
In contrast to the high-resolution and uniform sampling achieved by
digital cameras, the human (and many vertebrate) visual systems have
graded resolution, with a high-resolution fovea and decreasing
resolution in the periphery. This leads to behaviour where observers
make a sequence of fixations, with saccades between them to different
locations (see, e.g. \citealt*{findlay-gilchrist-03}). In Donald MacKay's
memorable phrase, vision is like a ``giant hand that samples the outside
world''.\footnote{Quoted on p.\ 23 of \cite{oregan-11}.}
Yet people's perception seems to be of a single, unified scene,
despite the large changes in retinal input that occur across saccades.
As \citet[sec.\ 1.4]{findlay-gilchrist-03} point out, this
fixation-saccade-fixation cycle leads to the questions: (i) where to
direct the gaze in order to take the next sample?  (ii) what
information is extracted during a fixation? (iii) how is the
information from one fixation integrated with that from previous and
subsequent fixations? These are the problems we tackle below.

The main inspiration for our work is the paper by
\citet{larochelle-hinton-10} (henceforth \landh) which tackles these
issues using a third-order Boltzmann machine model. In their paper the
Boltzmann machine has a set of hidden units $\bfz$, and a set of input units
for each fixation.  The hidden-to-fixation weights depend via a
third-order interaction on the location of the fixation. The
observations at each fixation are obtained by a ``retinal
transformation'' (RT), with high-resolution in the fovea and a
low-resolution periphery (see sec.\ \ref{sec:ret-transf} for more
details). We will sometimes refer to these fixations as \emph{glimpses}.
Their model is trained to reconstruct the glimpses, and also
to classify in input pattern (e.g., digit class for MNIST digits).

In the \landh\ model a high-resolution image can be synthesized after
having made several fixations, by making predictions at a dense grid
of locations. In our work we reformulate the task as to predict a
high-resolution image $\bfx$ from the latent variables $\bfz$, given
observations from several fixations.  This has the advantage that each
retinal transformation is then a \emph{known linear transformation} of
$\bfx$ based on the geometry, where a peripheral observation is the
(weighted) average of several high-resolution pixels. In contrast in
\landh\ the retinal transformation has to be \emph{learned} from data.

The task of choosing a sequence of fixations for a given scene can be
understood as maximizing the mutual information between $\bfz$ and the
observations. This task is known as Bayesian experimental
design (BED), and is discussed in sec.\ \ref{sec-oed}.  Below we
consider factor analysis (FA) and mixture of factor analyzers (MoFA)
models to relate $\bfz$ to $\bfx$. This has the advantage that the
linear retinal transformation combines nicely with the FA and MoFA
models to allow exact inference for the latent variables given the
observations. We demonstrate how these models can be used to predict
$\bfx$ given a sequence of observations, and also to learn the factor
analysis model for $\bfx$ from a set of glimpses of different input
images.

Our contributions are:
\begin{itemize}
\item Formulation of the task of fusing multiple glimpses in terms of
  a high-resolution latent image $\bfx$ and \emph{linear transformations} of
  this to yield the observed glimpses.
\item Use of the FA and MoFA models, which allow for exact inference
  of the latent variables, and learning of the models from data.
\item Formulation of ``where to look next'' as  an Bayesian
  experimental  design problem, and exact results  for BED for the
  FA model, and bounds for the MoFA model.
  \item These theoretical results are demonstrated on the Frey faces and
MNIST datasets.  
\end{itemize}  

The structure of the rest of the paper is as follows: in
sec.\ \ref{sec:methods} we discuss the retinal transformation, FA
and MoFA models, the fusion of multiple glimpses, Bayesian
experimental design, and the learning of models from RT data. Sec.\
\ref{sec:expts} describes experiments on the Frey faces and
MNIST datasets that illustrate BED and the learning of models from
foveal glimpses.
Sec. \ref{sec:relwork} discusses related work, and we conclude with
a discussion in sec.\ 5.

\section{Methods \label{sec:methods}}
\subsection{Retinal transformation \label{sec:ret-transf}}
\begin{figure}
  \begin{center}
  \begin{tabular}{cccc}  
    \raisebox{5mm}{\includegraphics[height=1.5in]{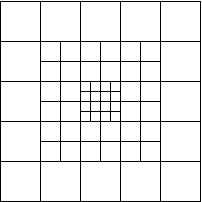}}
      & 
    \includegraphics[height=1.75in,width=1.3in]{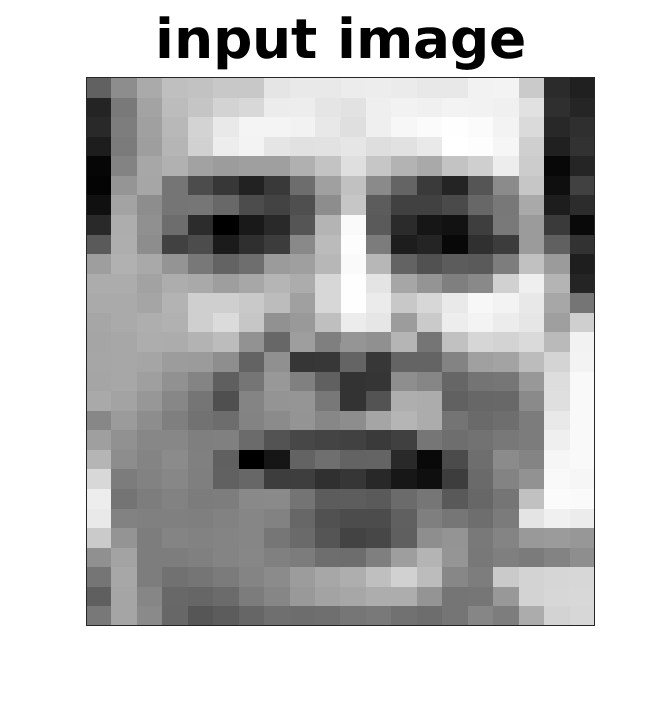} &
    \includegraphics[height=1.75in,width=1.3in]{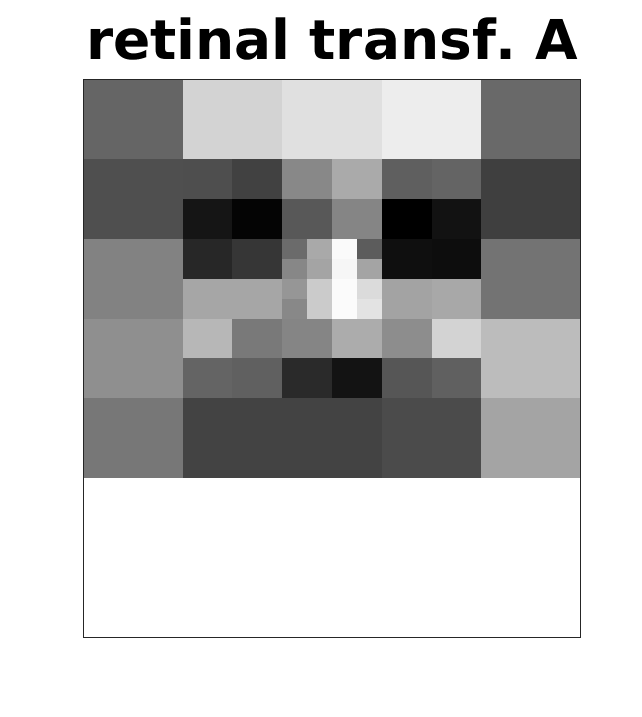} &
    \includegraphics[height=1.75in,width=1.3in]{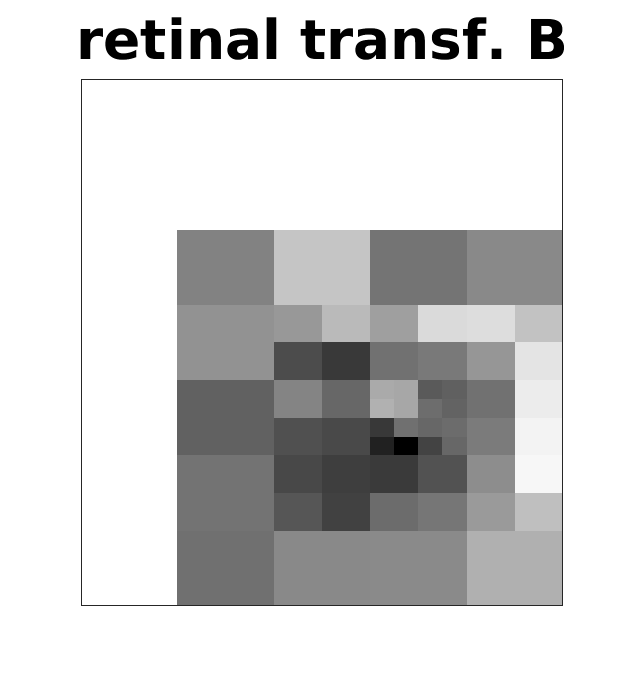} \\
    (a) & (b) & (c) & (d)
  \end{tabular}
  \end{center}  
  \caption{(a) The $20 \times 20$ retinal transformation used in our
    experiments. The innermost squares are $1 \times 1$ pixels, while
    the outermost are $4 \times 4$. (b) an image ($28 \times 20$) from the Frey
    dataset at  full resolution. (c) Visualization of the same
    image after the retinal transformation is applied to the top $20 \times 20$
    block of the image. Note: the bottom 8 rows of
    the original image are not observed, and are shown here as white.
    (d) Visualization of the input image under a different retinal
    transformation, with an offset $[8,4]$. Again regions of the input
    image that are not observed are shown as a white border.
    \label{fig-retinal-transf} }
\end{figure}

Let a high-resolution image $\bfx$ have $D$ pixels. A \emph{retinal
transformation} (RT) centered on location $\ell(a) = (r_a,
c_a)$\footnote{I.e. centered on row $r_a$ and column $c_a$.}
extracts high-resolution information only in the local neighbourhood
of $\ell(a)$, and lower-resolution (down-sampled) information in the
periphery, by averaging the values of pixels falling in each
peripheral receptive field. See Fig.\ \ref{fig-retinal-transf}(a)
for a visualization of the retinal transformation used in this work.
Fig.\ \ref{fig-retinal-transf}(c) visualizes the effect of this
transformation on the image shown in Fig.\ \ref{fig-retinal-transf}(b)
when the retinal transformation is applied to the top $20 \times 20$
block of the image. Fig.\ \ref{fig-retinal-transf}(d) illustrates a
different  RT obtained by shifting the centre 8 pixels down and 4 across.

The retinal transformation is a \emph{linear
transformation} of $\bfx$, and can be written as $\bfy_a = V_{\ell(a)}
\bfx$, where $\bfy_a$ is a vector of the intensities in each of the
cells in Fig.\ \ref{fig-retinal-transf}(a) centered at $\ell(a)$.
$\bfy_a$ has length $D^y_a < D$, and $V_{\ell(a)}$ is the
matrix that effects this transformation for location $\ell(a)$. Note
that the $V_{\ell(a)}$s are determined by the geometry, and need not
be learned. For some locations part of the retina will lie outside of
the image, and in this case those cells will return 0s in the relevant
part of $\bfy_a$.

\citet{larochelle-hinton-10} used a complicated arrangement of
hexagonal regions for their retinal transformation; in contrast we use
a simpler variable-resolution grid shown in
Fig.\ \ref{fig-retinal-transf}(a). However, in both cases the retinal
transformation is linear.  Linear down-sampling is used in the
super-resolution literature (see, e.g.,
\citealt*{chen-he-qing-wu-ren-sheriff-zhu-22}), but to our knowledge
this is always spatially uniform, as compared to the non-uniform
retinal transformation used here.

\subsection{Factor analysis and MoFA models}
For factor analysis we have
\begin{equation}
  \bfx = \bfmu + W \bfz + \bfe
\end{equation}  
where $\bfmu$ is the mean of $\bfx$, $\bfz \sim N(\bfzero,I_K)$ is a
standard multivariate Gaussian random variable of dimension $K$,
$W$ is the $D \times K$ factor loadings matrix,  and $\bfe$ is
a noise variable with $\bfe \sim N(\bfzero,\Psi)$, where $\Psi$ is a
diagonal matrix with non-negative entries.
Hence by integrating out $\bfz$ we have $\bfx \sim N(\bfmu, W W^T + \Psi)$.

Under the linear retinal transformation $\bfy_a = V_{\ell(a)} \bfx$, we again
have a factor analysis model $\bfy_a \sim N(\bfmu^y_a, V_{\ell(a)} W W^T V_{\ell(a)}^T
+ \Psi^y_{\ell(a)})$, where $\bfmu^y_a = V_{\ell(a)} \bfmu$.
Note here that we have \emph{not} transformed the observation noise
$\bfe$  from $\bfx$-space, but have instead assumed a FA model in
$\bfy$-space. Note also that in general $\Psi^y_{\ell(a)}$ can be
different for each location $\ell(a)$.

Standard Gaussian inference for $\bfz$ given $\bfy_a$ leads to
$\bfz|\bfy_a \sim N(\bfmu_{\bfz|\bfy_a}, \Sigma_{\bfz|\bfy_a})$ with
\begin{align}
  \Sigma^{-1}_{\bfz|\bfy_a} &= I_K + W_a^T (\Psi^y_{\ell(a)})^{-1} W_a,
  \label{eq:invpostcov} \\
  \bfmu_{\bfz|\bfy_a} &= \Sigma_{\bfz|\bfy_a} W_a^T 
  (\Psi^y_{\ell(a)})^{-1} (\bfy_a - \bfmu^y_a). \label{eq:postmean} 
\end{align}
where $W_a = V_{\ell(a)} W$, see, e.g., \citet[sec.\ 12.2.4]{bishop-06}.
The obvious estimator for reconstructing $\bfx$ given $\bfy_a$ is then
\begin{align}
  \hat{\bfx} = \bfmu +  W \bfmu_{\bfz|\bfy_a},
\end{align}
and one can also compute the
predicted covariance as $W \Sigma_{\bfz|\bfy_a} W^T + \Psi$.

A simple but powerful extension of the FA model is to use a mixture of factor
analyzers \citep{ghahramani-hinton-96}.
In $\bfx$-space we have $M$ components, with means, factor
loadings and noise variances$\{\bfmu^m, \; W^m, \; \Psi^m \}_{m=1}^M$,
and non-negative mixing proportions $\{ \pi^m \}_{m=1}^M$ which sum to
one. Each component indexed by $m$ has an associated latent variable $\bfz^m$.
Under the retinal transformation $\bfy_a = V_{\ell(a)} \bfx$, we have that
\begin{equation}
  p(\bfy_a) = \sum_{m=1}^M \pi_m p_m(\bfy_a) =
\sum_{m=1}^M \pi_m N(\bfy_a; \bfmu^{m}_a, W^m_a (W^m_a)^T + \Psi^{y,m}_{\ell(a)}) .
\label{eq:mix}
\end{equation}

One could also consider a (restricted) Boltzmann machine (RBM) model for
$\bfx$, as in \landh\ . For a RBM inference for the latents $\bfz$
from $\bfy$ observations is exact, but because of the partition
function, learning requires approximations; contrastive divergence
was used in \citet{larochelle-hinton-10}. In this paper
the (mixture of) FA model is used below, particularly as
it leads to exact results for the Bayesian experimental design problem.

\subsection{Fusing multiple glimpses}
Now assume that we have a sequence of $J$ observations
$\bfy_1, \ldots, \bfy_J$, 
at locations $\ell(1), \ldots, \ell(J)$. It is natural to write
\begin{equation}
p(\bfz, \bfy_{1:J}) = p(\bfz) \prod_{j=1}^J
p(\bfy_{j}|\bfz) . \label{eq:fusemult}
\end{equation}  
However, this is only correct if the $\bfy$'s provide \emph{disjoint}
information about $\bfx$. If there is overlap, this is not strictly
correct as it over-counts evidence.\footnote{This product rule is also
assumed, without comment, in eq.\ 6 of \citet{larochelle-hinton-10}.}

For the FA model, one can integrate out $\bfz$ from
eq.\ \ref{eq:fusemult} to yield $p(\bfy_{1:J})$. But as
everything is Gaussian, it is easiest to compute the mean and
covariance structure of the distribution for $\tilde{\bfy}$ which is
obtained by concatenating $\bfy_{1}, \ldots \bfy_{J}$.
Similarly we define $\tilde{\bfmu}$ by concatenating $\bfmu^y_{\ell(1)},
\ldots, \bfmu^y_{\ell(J)}$. $\tW$ is obtained by stacking the $W_j$s,
and  $\tilde{\Psi}^y$ is a $JD \times JD$ diagonal matrix with
$\Psi^y_{\ell(1)}, \ldots, \Psi^y_{\ell(J)}$ on the diagonal. Then we have that 
\begin{equation}
\tilde{\bfy} \sim N(\tilde{\bfmu}, \tW \tW^T + \tilde{\Psi}^y) , \label{eq:multobs}
\end{equation}
which is just a FA model for the extended vector $\tilde{\bfy}$. 

Eq.\ \ref{eq:fusemult} assumes that the latent state $\bfz$ is not
evolving over time. If it is, a natural extension is to use a linear
dynamical system (LDS) so that
\begin{equation}
p(\bfz_{1:J},\bfy_{1:J}) = p(\bfz_1) \prod_{j=2}^J
  p(\bfz_j|\bfz_{j-1}) \prod_{j=1}^J p(\bfy_j|\bfz_j), \label{eq:lds}
\end{equation}
where $p(\bfz_1)$ is $N(0,I_K)$, and $p(\bfz_j|\bfz_{j-1})$ is Gaussian.
For example one could use $\bfz_j = \alpha \bfz_{j-1} + \sqrt{1 -
  \alpha^2} \bfe_j$ for $0 \le \alpha \le 1$ and $\bfe_j \sim
N(\bfzero, I_K)$; the scaling of the noise is chosen to be variance
preserving under the unconditional $\bfz$ dynamics.
Use of the LDS model could also 
ameliorate the over-counting mentioned above, by creating some
``forgetting'' of older observations.

\subsection{Bayesian experimental design \label{sec-oed}}
The goal of experimental design for a given scene is to find the
sequence of fixations that optimize the amount of information they
provide about $\bfz$. We first briefly review the theory of
Bayesian experimental design as described, e.g., by
\citet{rainforth-foster-ivanova-bickford-smith-24}, but adapted to
our situation.

Consider an experimental design $\xi$, which in our case is the
location of the fixation. Given $\xi$ we obtain an observation
$\bfy_{\xi}$. The information gain about $\bfz$ given
$\bfy_{\xi}$ is defined as 
\begin{align}
  \mathrm{IG}(\bfz;\bfy_{\xi}) &= H[p(\bfz)] - H[p(\bfz|\bfy_{\xi})]
  = \bbE_{p(\bfz|\bfy_{\xi})} [\log
    p(\bfz|\bfy_{\xi})] - \bbE_{p(\bfz)} [ \log p(\bfz) ],
\end{align}
where $H$ denotes the (differential) entropy.
As $\bfy_{\xi}$ is unknown before a fixation, we target the
\emph{expected information gain} (EIG) which is given by
\begin{align}
  \mathrm{EIG}(\bfz|\xi) &= \bbE_{p(\bfy_{\xi})} \mathrm{IG}(\bfz;\bfy_{\xi}) \\
              &= \bbE_{p(\bfz) p(\bfy_{\xi}| \bfz)} [\log
    p(\bfz|\bfy_{\xi}) - \log p(\bfz) ],  \label{eq:eig2} \\
              &= \bbE_{p(\bfz) p(\bfy_{\xi}| \bfz)} [\log
    p(\bfy_{\xi}|\bfz) - \log p(\bfy_{\xi}) ].  \label{eq:eig3}
\end{align}
The EIG is equivalent to the mutual information between $\bfz$ and
$\bfy_{\xi}$, and the last line above is obtained from the one above via the two
ways of writing the mutual information $I(Y;Z) = H(Y) - H(Y|Z) = H(Z) - H(Z|Y)$.

To make a sequence of fixations, at each step we can consider the
\emph{incremental} EIG for a new fixation given the history. This is
known as an adaptive or sequential design, and the process as
Bayesian adaptive design (BAD).

We now apply these ideas to the FA model, using eq.\ \ref{eq:eig3}. 
The entropy of a multivariate Gaussian with covariance matrix
$\Sigma$ in $D$ dimensions is easily computed as
$\frac{D}{2} \log(2 \pi e) + \frac{1}{2} \log |\Sigma|$. The
expectation of the $- \log p(\bfy_{\xi})$ term is the entropy of
$p(\bfy_{\xi})$, which has covariance $W_{\xi} W^T_{\xi} + \Psi^y_{\xi}$.
The negative entropy coming from the $\log p(\bfy_{\xi}|\bfz)$ term
arises from the noise,
which has covariance $\Psi^y_{\xi}$. Plugging these entropies into eq.\
\ref{eq:eig3}, we obtain
\begin{equation}
\mathrm{EIG}(  \bfz | \xi) = \frac{1}{2} [ \log |W_{\xi} W_{\xi}^T + \Psi^y_{\xi}|
-  \log |\Psi^y_{\xi}| \; ]. \label{eig:y} 
\end{equation}

The above analysis can be readily extended to more than one
observation by replacing $\bfy_{\xi}$ with $\tilde{\bfy}_{\xi}$,
$W_{\xi}$ with $\tW_{\xi}$ and
$\Psi^y_{\xi}$ with $\tilde{\Psi}_{\xi}$, where these parameters are
defined near eq.\ \ref{eq:multobs}.
For say $J=2$ it is quite reasonable to do the search over
all combinations, but for larger $J$ it would make sense to do this
greedily. 

Optimal experimental design for the mixture of FA model is derived 
in Appendix \ref{sec:oed-mofa}, and leads to the result
\begin{equation} 
\mathrm{EIG(mix)} \le  H(\pi) + \sum_{m=1}^M \pi_m
\mathrm{EIG}_m(\bfz_m|\xi) , \label{eq:eigUB}
\end{equation}
where $H(\pi)$ is the entropy of the mixing proportions.
This upper  bound is tight when the Gaussians are well separated. 
The bound on
$\mathrm{EIG(mix)}$ is basically a weighted average of the individual
$\mathrm{EIG}$'s for each Gaussian component, plus $H(\pi)$.
One could also make use of a lower bound on the entropy of a mixture,
as given in Theorem 2 of
\citet{huber-bailey-durrant-whyte-hanebeck-08}.  However, in our
experiments (see below) we have found the gap between the upper and lower bound is
large for our data, and given that the mixture components are quite well
separated (as judged by the posterior probabilities of datapoints) we
prefer the upper bound.

The property of a Gaussian that the posterior covariance is
independent of the particular value observed for $\bfy_{\xi}$,
but only on the design $\xi$,  means that the optimal design for
the FA model (and for the MoFA upper bound) can be determined before
test data is observed, and is thus not an adaptive design. But with
more complex models (see sec.\ \ref{sec:discuss}) optimization of the
EIG criterion would likely lead to adaptive designs.

The EIG criterion is a generic criterion, aimed at maximizing the
amount of information that the observations provide about $\bfz$.
This can be contrasted with \emph{task-specific} strategies.
Famously, \citet{yarbus-67} observed \emph{different} patterns of
fixations when giving subjects different task instructions \emph{when
observing the same image}. See also \citet{hayhoe-ballard-05} for a
review of more recent work on this topic. \landh\ did not do BED, but
instead made use of classification labels and trained a controller to
assign high scores to fixation positions which were more likely to
make a correct prediction of the true label. Similar criteria have
been used by later workers, e.g.,
\citet*{mnih-heess-graves-kavukcuoglu-14}.

To our knowledge the use of EIG for determining fixation locations is
novel, as are the bounds for the MI between $\bfy$-data and the latent
representation for a mixture of factor analyzers.  Other work on the
next-best-view problem and information-theoretic criteria for
directing saccades is discussed in sec.\ \ref{sec:relwork}.

\subsection{Learning $W$ and the $\Psi^{y}_i$s \label{sec:learn}}
For the FA model for $\bfx$, there is not a closed-form solution for
the maximum likelihood parameters for $W$ and $\Psi$ given data
samples $\bfx^1, \ldots, \bfx^N$. \citet[ch.\ 9]{mardia-kent-bibby-79}
describe the principal factor analysis and maximum likelihood methods
for estimating the parameters. Given an estimate of $W$, $\Psi$ can be
estimated as $\mathrm{diag}(C_x - W W^T)$, where $C_x$ is the
covariance of $\bfx$, assuming that all of the resulting entries are
positive.  Another standard approach for estimating the parameters is
to use the EM algorithm, see, e.g., \citet{rubin-thayer-82}. In
contrast, for the Probabilistic PCA model of \citet{tipping-bishop-99}
there is a closed form solution based on the eigendecomposition of the
covariance matrix of the data.

In the case of foveal glimpses, we have data $Y = (\bfy^1, \ldots,
\bfy^n)$, where each $\bfy^i$ has an associated retinal transformation
$V_{\ell(i)}$. In our experiments we generate $\bfy^i$s by first
choosing an $\bfx$ sample randomly, and then choosing a random retinal
location.  This is repeated $n$ times.

The additional complication of the $V_{\ell(i)}$s means
that we were  not able to derive an EM algorithm to estimate
$W$ and the $\Psi^{y}_{\ell(i)}$s. The log likelihood for $Y$ is given by 
\begin{equation}
L = - \frac{1}{2} \sum_{i=1}^n (\bfy^i)^T (V_{\ell(i)} W W^T
V^T_{\ell(i)} + \Psi^y_{\ell(i)})^{-1} \bfy^i - \frac{1}{2}
\sum_{i=1}^n \log |V_{\ell(i)} W W^T V^T_{\ell(i)} + \Psi^y_{\ell(i)}|
+ c \label{eq:ll}
\end{equation}
where $c$ is a constant independent of the parameters of interest.
It is also possible to write down the log likelihood when there
  are multiple RTs for each image, using the set-up as in
  eq.\ \ref{eq:multobs} with the concatenated observations
  $\tilde{\bfy}^i$ for image $i$, $\tilde{V}_i$ being the stacked
  retinal transformations, and $\tilde{\Psi}^y_i$ being the block
  diagonal matrix of noise variances.
As described in Appendix \ref{sec:dLL} we can obtain derivatives of $L$
with respect to $W$ and the $\Psi^{y}_i$s, and use gradient ascent to
optimize it. 

To initialize $W$, we apply PPCA (a special case of FA) to data
upsampled from the retinal transformation to $\bfx$. For example,
if one cell in $\bfy$ is obtained by averaging several pixels in
$\bfx$, then a crude version of $\bfx$ can be obtained by giving each
of the pixels  the same (averaged) value obtained from $\bfy$. As the
retinal transformation does not get input from the whole of $\bfx$,
the unobserved pixels are treated as missing, using the variational
Bayesian algorithm PCAMV from \citet{ilin-raiko-10}. This also returns
the estimated PPCA noise variance, which can be used as a guess
for $\Psi^x$. The individual $\Psi^y_{\ell(i)}$ matrices can then
be initialized as $\mathrm{diag}(V_{\ell(i)} \Psi^x V^T_{\ell(i)})$.

To learn the parameters of the mixture model (eq.\ \ref{eq:mix}) we
optimize the log likelihood $L_{mix} = \sum_i \log p(\bfy^i)$.
Consider a parameter $\theta_c$ that belongs to mixture component $c$.
We then have (after some manipulation)
\begin{equation}
  \frac{\partial L_{mix}}{\partial \theta_c} = \sum_{i=1}^n p(c|\bfy^i)
  \frac{\partial \log p_c(\bfy^i)}{\partial \theta_c} .
\end{equation}
The last factor $\partial \log p_c(\bfy^i) / \partial \theta_c$ is
exactly what has been computed in Appendix \ref{sec:dLL}. Hence we can
use gradient-based optimization for the mixture parameters
$\{\pi^m, \; \bfmu^m, \; W^m, \; \Psi^m \}_{m=1}^M$.

\section{Experiments \label{sec:expts}}
We carry out experiments on the Frey faces dataset and the MNIST
digits dataset, described below. \matlab\ code used for the
experiments is available.\footnote{Code available at
\url{https://homepages.inf.ed.ac.uk/ckiw/mypages/software.html}.}

{\bf Frey faces dataset:}\footnote{
available from e.g., \url{https://github.com/SheffieldML/GPmat/blob/master/datasets/data/frey_rawface.mat}.},
This consists of 1965 frames of a greyscale
video sequence with resolution $20 \times 28$ pixels. Pixel intensities were
rescaled to lie between -1 and 1. For the experiments reported in
sec.\ \ref{sec:exp-frey-oed} the data was split 80:20 into a training and
testing set. Carrying out PCA on the data indicated that 43 components
explained over 90\% of the data variability; thus PPCA and FA models
were fitted using 43 latent components. 

The retinal transformation used was the $20 \times 20$ variable
resolution grid shown in Fig.\ \ref{fig-retinal-transf}(a). The
``home'' position for the grid was chosen to occupy the top
$20 \times 20$ block of the image, as illustrated in
Fig.\ \ref{fig-retinal-transf}(c). Horizontal offsets of
$[-8, \; -4, \; 0, \; 4, \; 8]$ pixels were used, and vertical offsets
of $[-8, \; -4, \; 0, \; 4, \; 8 \; 12, \; 16]$ pixels. The gives
$5 \times 7 = 35$ different possible offsets.
These ranges were chosen so as to move the fovea to all corners
  of the image.
With some offsets, some cells of RT will lie outside the image, and thus receive
no input. To handle inference and learning in this situation we make
computations with $\bfy_a$ and the corresponding matrices using only
the active entries, i.e. the ones that do receive input.

{\bf MNIST dataset:}\footnote{Data and loading functions were obtained
from \url{https://github.com/mkisantal/matlab-mnist/blob/master/}.}
The full dataset contains 50,000 training and 10,000 test images each
of size $28 \times 28$, but for the experiments here were used only
examples of the digit 2, with 5,958 training and 1,032 test examples.

A 10 component PPCA model was fitted to the 2s data using Ian Nabney's Netlab
\texttt{gmm}
software.\footnote{\url{https://github.com/sods/netlab}.}
This initially uses k-means, and then fits a PPCA model to the data
belonging to each component. Using 70 latent dimensions in each
component explained over 90\% of the data variability in 8 out of the
10 components (and was close on the other two).
The retinal transformation was the same $20 \times 20$ variable
resolution grid described above, but now both the horizontal and
vertical offsets were both set to $[-4, \; 0, \; 4, \; 8, \; 12, \;
  16]$ pixels, giving $6 \times 6 = 36$ possible offsets. The handling
of cells of the RT lying outside the image was as described above.

\begin{table}
  \begin{tabular}{cc}
       \hspace*{-5mm}    
       {\small     
       \begin{tabular}{lcccc}
       \multicolumn{5}{c}{RMSE error} \\ \hline
       Design & 0 & 1 & 2 & FA \\ \hline
       $W_{FA}$ BED & 0.2097  &  0.1126  &   0.0952 &   0.0790       \\
       $W_{FA}$ Random   &  " &   0.1251 &   0.1081 &  " \\ \hline
       $W_{optY}$ BED     &  " &   0.1454 &   0.1282 &   0.1038 \\
       $W_{optY}$ Random  &  " &   0.1552 &   0.1490 &   "
       \end{tabular}
       \hspace*{10mm}           
       \begin{tabular}{ll}
        \multicolumn{2}{c}{Log likelihoods}  \\ \hline 
        Method & LL/ex.\ \\ \hline 
        Independent & 69.08 \\
        PCAMV soln ($\Psi^y$s opt) & 90.95 \\        
        FA soln ($\Psi^y$s opt)   & 107.52 \\
        Opt from PCAMV & 115.59 \\
      \end{tabular}
    } 
    \end{tabular}
    \vspace*{3mm}
    \caption{Frey faces data: (Left) RMSE error on the test set for 0,
      1 and 2 fixations, for the BED and random designs for both
      $W_{FA}$ and $W_{optY}$. The last column, marked FA, is the
      reconstruction error that can be achieved using the whole image
      $\bfx$ as input. (Right) Table showing the log likelihood per
      training example for 4 different models when learning the
      parameters.
    \label{tab:design}}
\end{table}

\begin{figure}
  \begin{center}
  \begin{tabular}{ccccc}  
    \includegraphics[height=1.4in,width=1.0in]{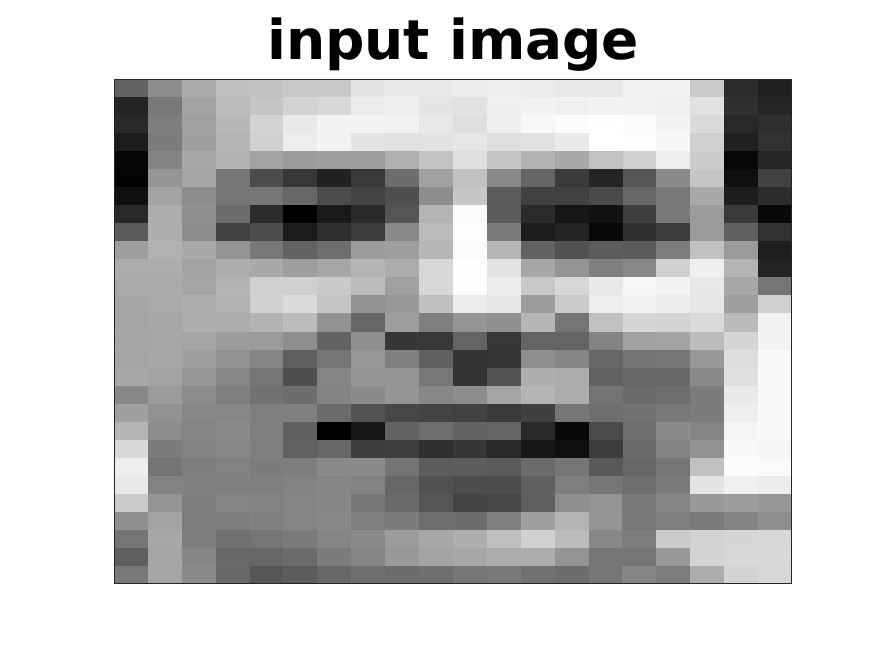} &
    \includegraphics[height=1.4in,width=1.0in]{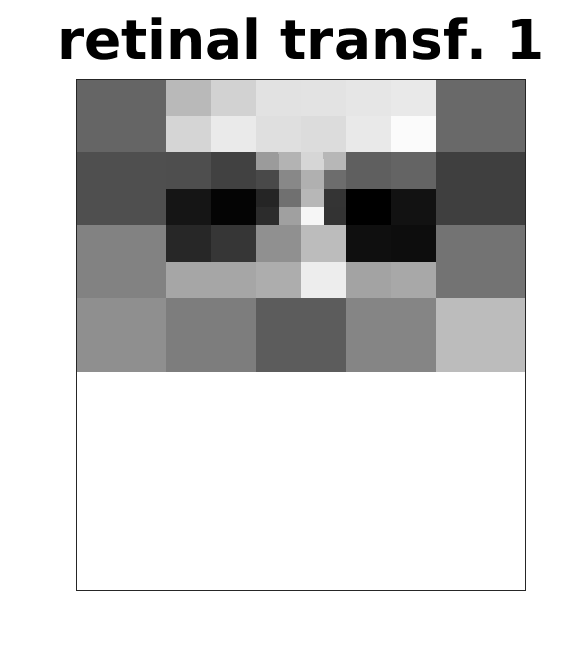} &
    \includegraphics[height=1.4in,width=1.0in]{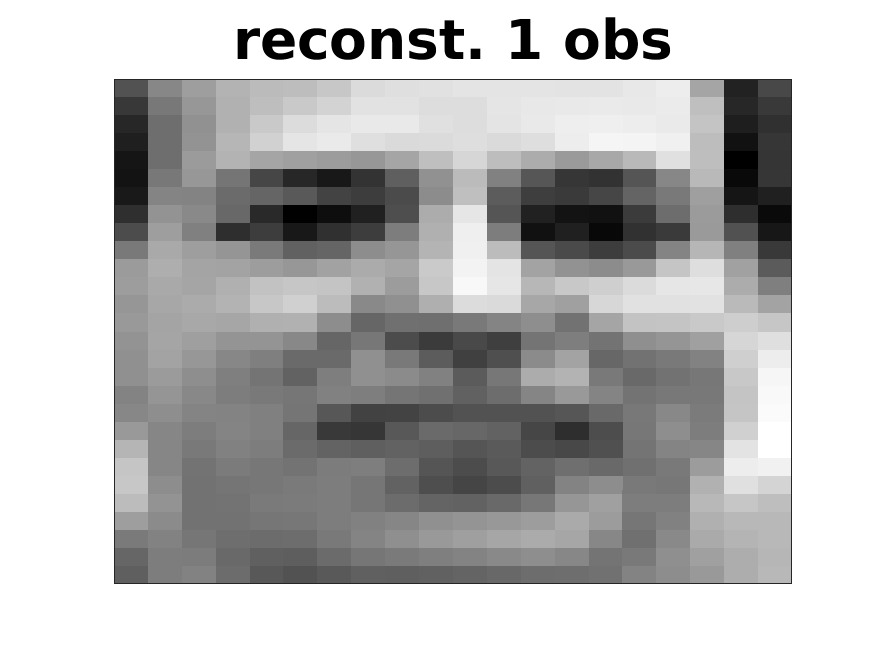} &
    \includegraphics[height=1.4in,width=1.0in]{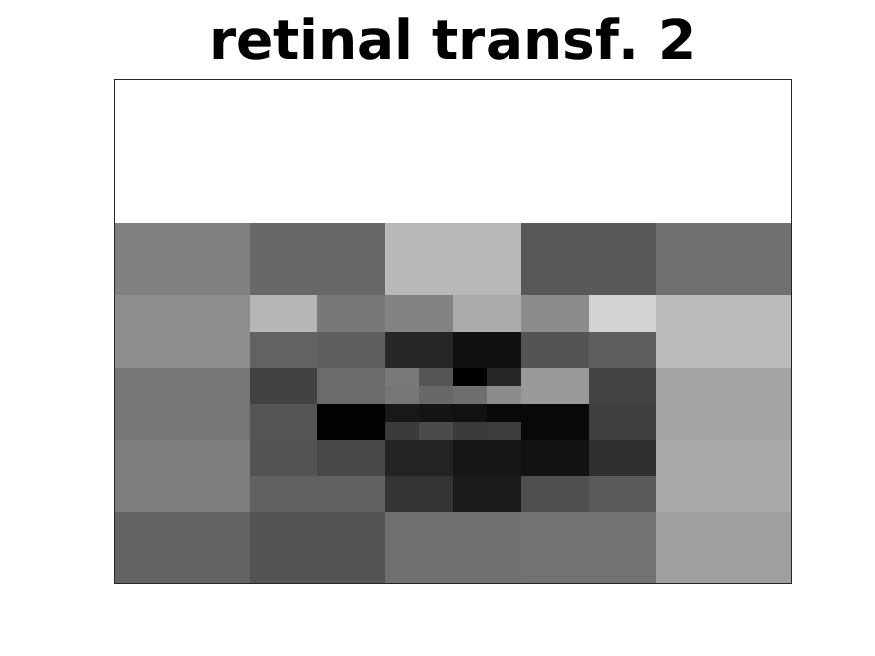} &
    \includegraphics[height=1.4in,width=1.0in]{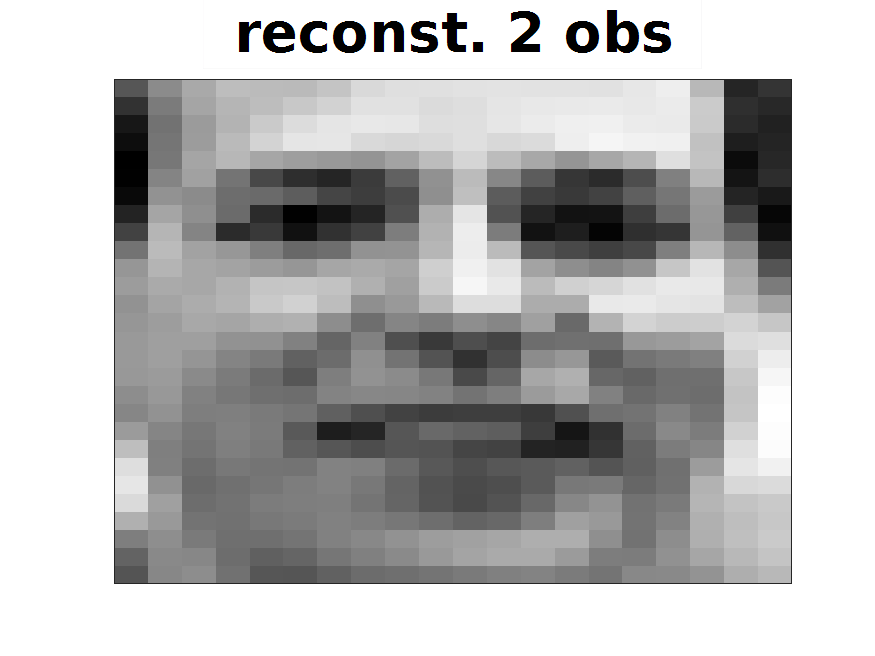} \\
    (a) & (b) & (c) & (d) & (e)
  \end{tabular}
  \end{center}
  \caption{(a) Original Frey face image, (b) retinal transformation 1,
    (c) reconstruction from this RT, (d) retinal transformation 2,
    (e) reconstruction from both RTs. \label{fig:freyRT1RT2}
   } 
\end{figure}

\subsection{Frey faces: Experimental design \label{sec:exp-frey-oed}}
For this set of experiments $W$ was determined 
using the \texttt{factoran} function in \matlab\  on the training
$\bfx$-data. For each offset index by $a$, we have that
$W_a = V_{\ell(a)} W$ and the corresponding $\Psi^y_{\ell(a)}$ was estimated as
explained in the first paragraph of sec.\ \ref{sec:learn}.

As shown in sec.\ \ref{sec-oed}, maximizing the expected information gain
is achieved by minimizing $\bbE_{\bfy_{\xi}} H[p(\bfz|\bfy_{\xi})]$. 
Searching over pairs of offsets, the optimum offsets are obtained
as $o_1 = [-4, \; 0]$ and $o_2 = [8, \; 0]$, where the vertical
offset is given first, then the horizontal one. So $o_1$ and $o_2$ shift the
grid 4 pixels up and 8 pixels down relative to
Fig.\ \ref{fig-retinal-transf}(c),
as shown in Figs.\ \ref{fig:freyRT1RT2}(b) and (d).
For comparison purposes, we use a random
design, where for each input image, a random pair of offsets are
chosen. For a given input image, we first reconstruct it based on the
fixation at $o_1$, or at a random offset. We then add either
an observation at $o_2$ (for the
optimal design), or a second random offset (for the random design).
An example of reconstructions based on either $o_1$ or both $o_1$ and $o_2$
are shown in Figs.\ \ref{fig:freyRT1RT2}(c) and (e). RT1 focuses on
the top of the image, so the mouth is less well reconstructed, but
this is improved in panel (e) after the use of RT2 as well.

Table \ref{tab:design}(left) top two rows shows the RMSE error on the
test set as a function of the number of fixations. For 0 observations
we simply use the overall mean image to reconstruct the input. The
right-hand entry (marked FA) is the reconstruction that can be
achieved using the whole image $\bfx$ as input. This gives a lower
bound on what can be achieved. The RMSEs for the 0 and FA fixations
are the same for both designs. For 1 and 2 fixations, the RMSE is (as
expected) lower for the optimal design compared to the random design,
indeed on average the RMSE from one optimal fixation is close to that 
from two random fixations.  One can make a \emph{paired comparison} of
the error on each image for the optimal and random designs. For 1
fixation, 276 out of 393 differences were in favour of the optimal
design (p-value $7.65 \times 10^{-17}$ according to the sign test),
and for 2 fixations 320 out of 393 differences (p-value $2.33 \times
10^{-35}$).  This shows conclusively (as expected) that the optimal
design is superior to a random design.

\subsection{Frey faces: Learning the parameters from $Y$ data}
Above the $W$ matrix (call this $W_{FA}$) was estimated using FA
on the high-resolution $\bfx$
data. We now show that a result of similar quality can be obtained
based on the RT data (the $\bfy$'s). For each of the 35 possible
offsets, 100 examples were chosen randomly from the Frey faces data
and the corresponding RT obtained. Because each RT does not
cover the whole image, 55\% of the entries
in the upsampled images (see sec.\ \ref{sec:learn}) were missing.
The variational Bayesian algorithm of
\citet{ilin-raiko-10} can handle this missing data,
and was used to create an initial PPCA solution.\footnote{Code
available at \url{https://users.ics.aalto.fi/alexilin/software/}.}
We then used scaled conjugate gradient (SCG) search \citep{moller-93} to
optimize the log likelihood in eq.\ \ref{eq:ll}.

Table \ref{tab:design}(right) shows the log likelihood (LL)
per training example for a number of different models.  As a baseline, the $Y$
data is modelled using an independent Gaussian for each dimension
(estimated separately for each offset). This gives a LL of 69.08.
Using the PPCA solution obtained from the PCAMV algorithm
and optimizing only the $\{ \Psi^y_{\ell(a)} \}$ matrices gives 90.95,
and optimizing both $W$ and the $\{ \Psi^y_{\ell(a)} \}$s gives a LL of 115.59.
For a comparison, if we fix
$W_{FA}$ and optimize only the $\{ \Psi^y_{\ell(a)} \}$ matrices, we
obtain a LL of 107.52. Optimizing both $W$ and the $\{ \Psi^y_{\ell(a)} \}$
matrices from the FA solution gives essentially the same LL as from the PCAMV
initialization. Although the absolute values of the log likelihood are
not very meaningful, the relative differences to the baseline are.
The fact that a better LL can be obtained by optimizing $W$ relative
to the FA solution tells us that $W_{FA}$ is not the optimal solution
for the $Y$ data, but the LL gap between them is not very large.

We can also evaluate the $W$ found above by optimizing on the $Y$-data
from the PCAMV initialization (call this $W_{optY}$) in terms of the
reconstruction task from sec.\ \ref{sec:exp-frey-oed} above.
The results are shown in the lower rows of Table
\ref{tab:design}(left). The reconstruction errors using the optimal
designs are a bit larger for $W_{optY}$ than for $W_{FA}$,
but they are still very effective.  Note that the lower
bound reconstruction error obtained by using the full $\bfx$ input has
also increased relative to the FA solution, as $W_{optY}$ is suboptimal
relative to $W_{FA}$.

\begin{figure}
  \begin{center}
  \begin{tabular}{cccc}
    \hspace*{-3mm}
    \includegraphics[height=1.3in,width=1.3in]{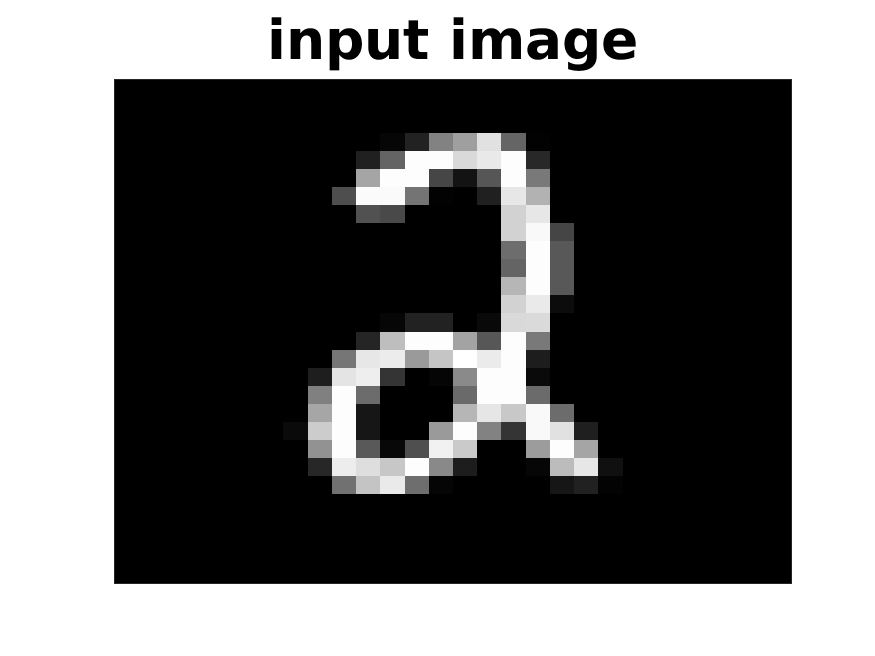}    & 
    \includegraphics[height=1.3in,width=1.3in]{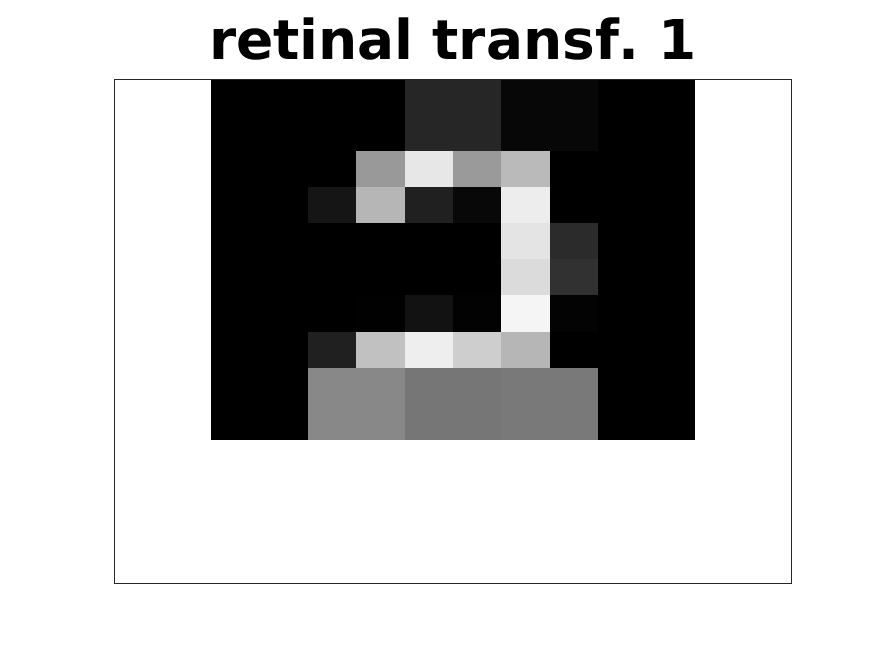} &
    \includegraphics[height=1.3in,width=1.3in]{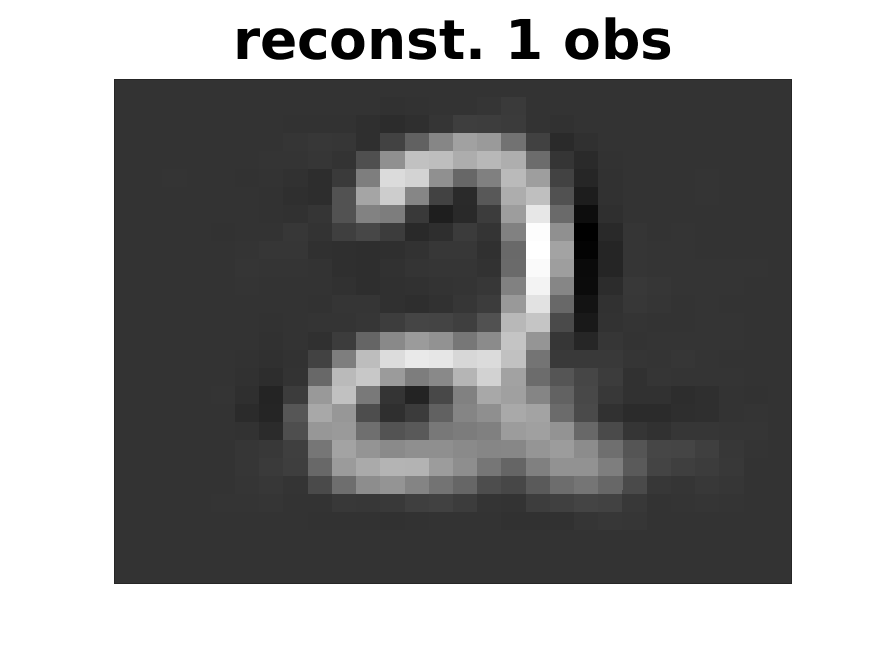} &
    \includegraphics[height=1.3in,width=1.3in]{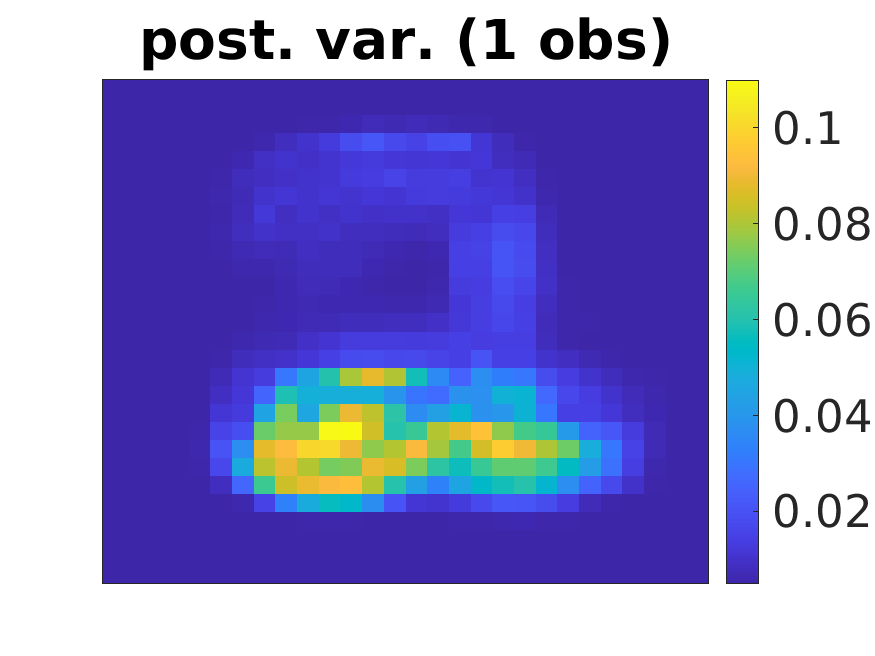} \\
    \hspace*{-3mm}
    \includegraphics[height=1.3in,width=1.3in]{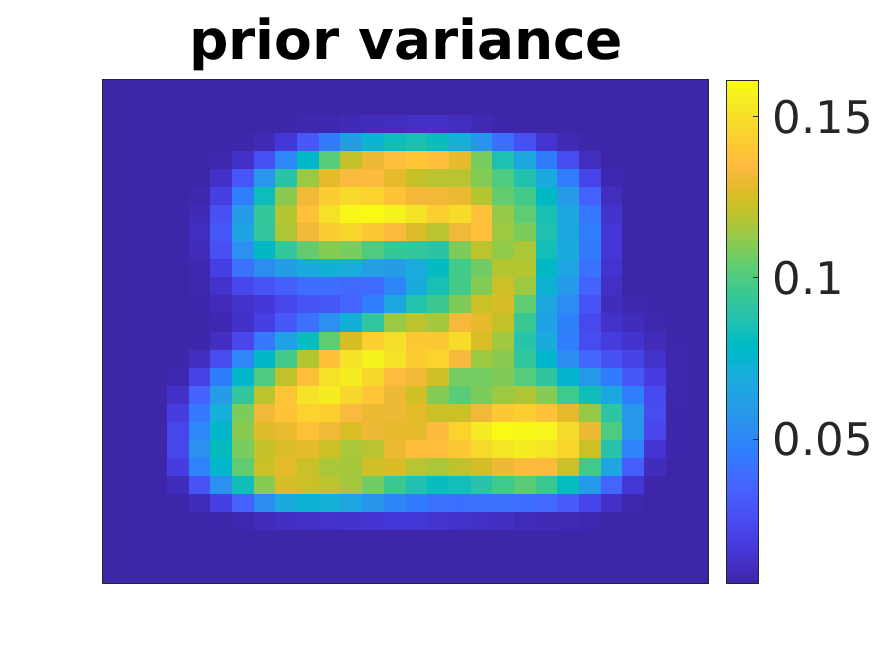}    & 
    \includegraphics[height=1.3in,width=1.3in]{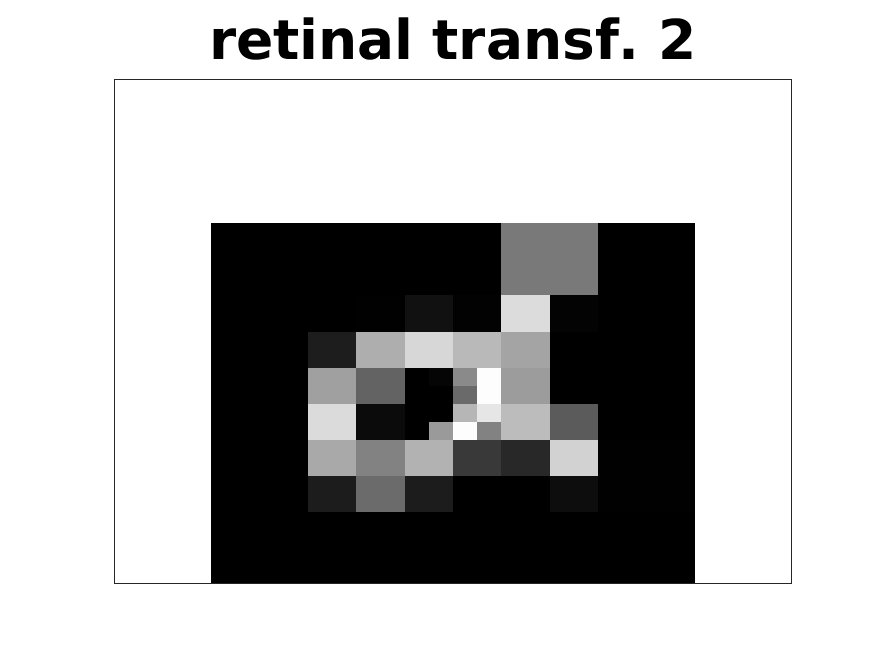} &
    \includegraphics[height=1.3in,width=1.3in]{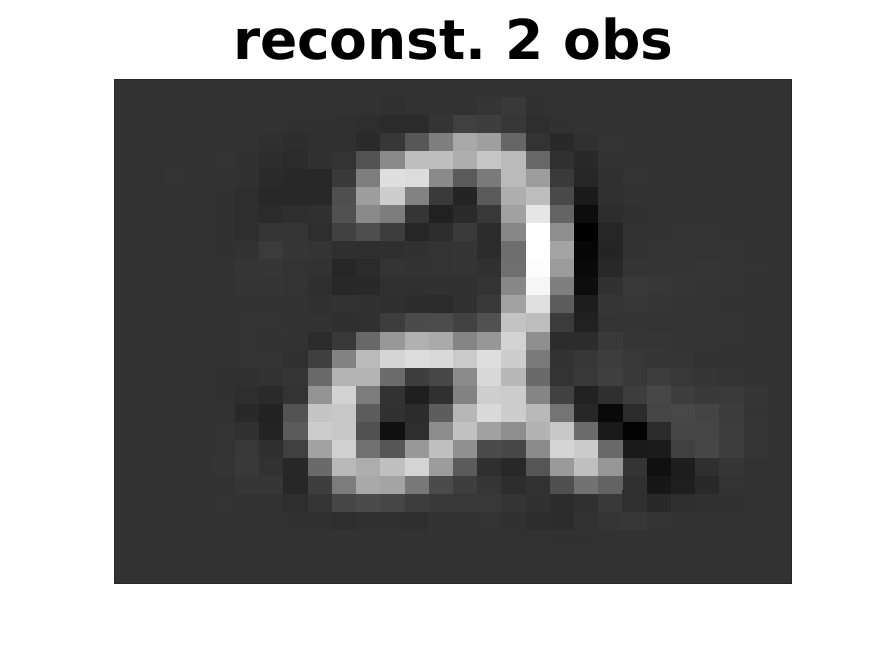}    & 
    \includegraphics[height=1.3in,width=1.3in]{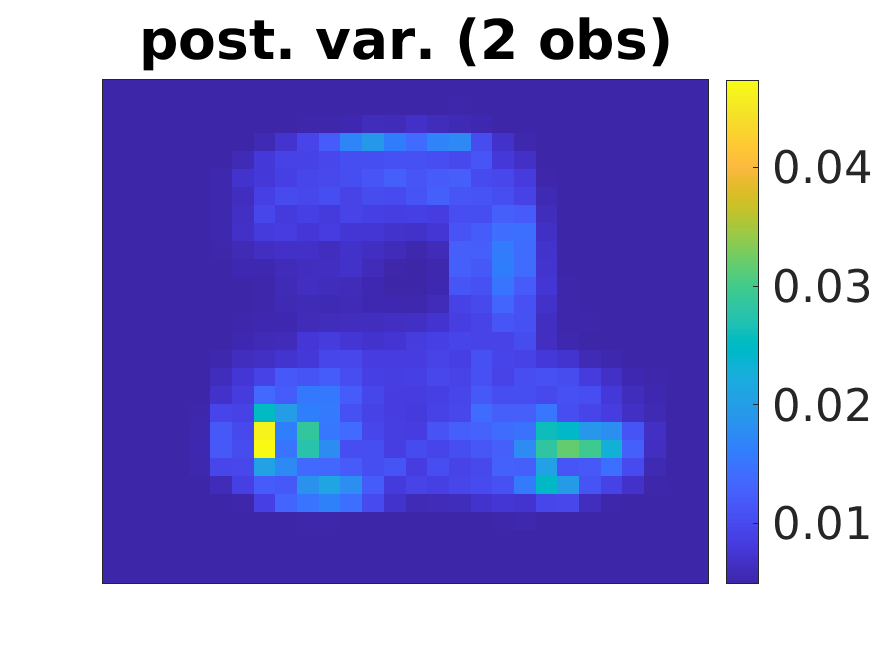}
  \end{tabular}
  \end{center}
  \caption{Top row: An example image (left) undergoes RT1
    and is reconstructed as in the third panel.
    The posterior variance after RT1 (4th panel)  is high towards the
    bottom of the image where there were no observations.
    Bottom row: the prior variance (averaged over the 10 components) is shown
    on the left.  A second retinal transformation is applied, and the resulting
    reconstruction using both RTs is shown in the third panel. The posterior
    variance after RT1 and RT2 is much reduced relative to RT1 only.
    \label{fig:MNIST}
   } 
\end{figure}

\subsection{MNIST 2s: Experimental design \label{sec:exp-mnist-oed}}
For this experiment the means, factor loadings and noise variances
$\{\bfmu^m, \; W^m, \; \Psi^m \}_{m=1}^M$ were obtained in
$\bfx$-space using the Netlab \texttt{gmm} functionality for PPCA
\citep{nabney-02} on the training data.  For each offset indexed by
$a$, we have that $W^m_a = V_{\ell(a)} W^m$ and $\bfmu^m_a =
V_{\ell(a)} \bfmu^m$, and the corresponding $\Psi^{y,m}_{\ell(a)}$s
were estimated by gradient ascent on $L_{mix}$.  A search over pairs
of offsets determined that $o_1 = [0,4]$ and $o_2 = [8,4]$ were
optimal with respect to the EIG upper bound of eq.\ \ref{eq:eigUB}. So
$o_1$ focuses centrally on the top of the image, and $o_2$ 
centrally on the bottom.
The retinal transformations corresponding to $o_1$ and $o_2$ are
shown in the second panels (top and bottom) of Fig.\ \ref{fig:MNIST}.

For a given test example one can compute the posterior
distribution over the components, and the entropy of this
distribution. Most test points have entropy near zero (they are
associated with just one component), but some are associated with two
or more components. For the mixture model in $\bfx$-space, only
23 out of 1032 test points have an entropy of more than 0.0808 bits
(corresponding to non-zero probabilities of 0.99 and 0.01).
Using only one fixation (at $o_1$), there are 165 test points  with
entropy above this threshold, but this drops to 102 with two
fixations (at $o_1$ and
$o_2$). (As expected, providing more information makes the
posteriors more concentrated.) The EIG upper bound is tight
when the components are well separated, and the relatively low
entropy of the posteriors suggests that this is usually the case.

Fig.\ \ref{fig:MNIST} shows the reconstruction of an input image (top
left) using either RT1 or both RT1 and RT2. The prior variance of each
pixel (bottom left) is obtained by averaging the variances coming from
each component by the mixing proportions. After observing RT1, the
posterior is heavily concentrated on one component. As RT1 focuses towards
the top of the image, the posterior variance (top row, rightmost
panel) is larger towards the bottom and the reconstruction is
blurry here. But after RT2 the posterior variance is reduced and the
reconstruction is sharper.

\begin{table}
  \begin{center}
      \begin{tabular}{lcccc}
       \multicolumn{5}{c}{RMSE error} \\ \hline
       Design & 0 & 1 & 2 & FA \\ \hline
       BED & 0.2525 & 0.1416  &  0.1078 &   0.0788   \\
       Random   &  " &   0.2058 &   0.1734  &  " 
      \end{tabular}
   \end{center}   
  \caption{MNIST 2s data: RMSE error  on the test set
      for 0, 1 and 2 fixations, for the BED and
      random designs. The last column, marked FA, is the reconstruction
      error that can be achieved using the whole image $\bfx$ as input. 
    \label{tab:mnist}}
\end{table}

As with the Frey data, one can compute the reconstruction error
given zero, one or two observations. For the mixture model this is
a little more complex as we have that
$p(\bfx|\bfy_{\xi}) = \sum_m p(\bfx|c=m, \bfy_{\xi})
p(c=m|\bfy_{\xi})$. To compute $\bbE[\bfx|\bfy_{\xi}]$ we use 
$\bbE[\bfx|c=m, \bfy_{\xi}] = \bfmu^m + W^m
\bbE[\bfz^m|\bfy_{\xi}]$, so the prediction $\bbE[\bfx|\bfy_{\xi}]$
is a weighted average of the predictions from each component.
(Of course it would also be possible to use a probabilistic prediction
using the full mixture model.) The results are shown in Table
\ref{tab:mnist} and follow a similar pattern as for the
Frey faces data, i.e.\
for 1 and 2 fixations, the RMSE is (as expected) lower for the optimal
design compared to the random design, and indeed on average the RMSE from
one optimal fixation is better than from two random fixations. Paired
comparisons of the error on each image show that for 1 fixation,
884/1032 differences were in favour of the optimal design (sign test p-value
$4.92 \times 10^{-129}$) and for 2 fixations, it was
993/1032 (p-value $2.13 \times 10^{-193}$).

\section{Related work \label{sec:relwork}}
In the introduction we have discussed \citet{larochelle-hinton-10} and
how it relates to our work. Further related work is covered below,
under various headings.

\paragraph{3D reconstruction from multiple views:} The problem of
novel view synthesis (NVS) is to use images taken from a number of points
of view of a scene to synthesize an image of a  novel view of that
scene. This shares with our problem the task of fusing
multiple views, although with standard cameras it does not have to
handle variable resolution retinal transforms.

The generative query network (GQN) of \citet{eslami-gqn-18} fuses
multiple 3D views in an unsupervised manner to allow NVS.  In this it
faces a similar (but more general) task to our model, but without
retinal transforms. The authors do not use BED to select new
viewpoints, although they do compute a ``predicted information gain''
measure (equivalent to EIG) which would have allowed them to do this.

More recently the Neural Radiance Field (NeRF) model of
\citet{mildenhall-srinivasan-tancik-barron-ramamoorthi-ng-20} and
subsequent developments has become popular for this 3D task. It
builds in more geometric structure than the GQN.  The
original NeRF model is differentiable (facilitating learning) and can
be used to predict novel views given multi-view data for a single
scene. However, \citet{jang-agapito-21} generalized this to include
the shape and appearance latent variables, making it a closer
match to our work. However, again these authors do not address issues of the
retinal transformations or BED.

Within the domain of active 3D object reconstruction, the
question of view planning arises naturally, and is known as the
\emph{Next-Best-View} (NBV) problem. For methods which use a
volumetric representation (e.g., voxels), \emph{volumetric information gain}
(VI) criteria are commonly used. For example \citet{kriegel-etal-15}
compute a binary probability for each voxel of whether it is occupied
or not. A new viewpoint is selected based on
maximizing the average entropy of voxels belonging to rays
projected from the viewpoint.. More complex VI criteria are proposed
in \citet{delmerico-etal-18}. With the recent rise of NeRFs,
evaluation of the uncertainty in the NeRF representations has also been
used for NBV planning
\citep{lee-chen-wang-liniger-kumar-yu-22,ran-zeng-he-li--chen-lee-chen-ye-23}.
The above references are tackling a rather different problem to our work, as
they mainly focus on volumetric uncertainty. Also,
in contrast with our use of EIG, they lack a latent
$\bfz$, and evaluate the uncertainty by averaging it over voxels; 
for the foveal fixations problem this is rather like
having an independent prior for each pixel.

\paragraph{Saliency maps:} A popular approach to determining candidate
fixation locations is through a \emph{saliency map} which is computed
bottom-up from image features. A classic work is by
\citet{itti-koch-01} who used colour, intensity and orientation
features as inputs. More recently deep learning has been applied to
predicting saliency maps with greater accuracy, see e.g.,
\citet{kuemmerer-wallis-bethge-16}.  However, such bottom-up
approaches do not address the same problem as selecting fixations in
order to maximize information about $\bfz$. Note also that computing
a saliency map from a high-resolution image misses the point that
the eye has graded resolution and does not have this panoramic view
available; instead it must decide where to look next based on the
history of the fixation positions and what was observed at each
fixation.

\paragraph{Classifiers:}
\cite{mnih-heess-graves-kavukcuoglu-14}	used a Recurrent Attention
Model (RAM) to combine multiple	glimpses for a classification task.  A
glimpse sensor takes patches of various resolutions centered at a given
location $\ell_t$, and uses them to update a hidden state $h_t$, which
depends on $h_{t-1}$ and the current glimpse. $h_t$ is then used to
predict where to look next (i.e., $\ell_{t+1}$) and also to make a
prediction for a class label. Note that	this work only provides
a classifier, and not a reconstruction of the input image.
Recently \citet{zoran-chrzanowski-huang-gowal-mott-kohli-20} have used
a more modern architecture, adding a visual attention component	
guided by a recurrent (LSTM) top-down sequential process
to a ResNet architecture. However, the attention map does not carry
out fixations, and in fact is multiplied pointwise with	a values
tensor, then summed across the spatial dimension in order to
feed into the LSTM. The	LSTM state is then used	to predict the
classification (and there is no reconstruction of the input).

\paragraph{Psychology literature:}
\citet[p.\ 323]{hochberg-68} postulated that trans-saccadic
integration takes place via a \emph{schematic map}, by which he meant
``the program of possible samplings of an extended scene, and of
contingent expectancies of what will be seen as a result of those
samplings''. Our interpretation of the last part of this sentence is
that predictions can be made, based on the fixations so far, as to
what will be seen when the eyes are moved.

\citet[p.\ 313]{mackay-73} makes a similar point, that the aim of the
observer is to build up an internal representation (our $\bfz$) of
the world. In a static world, once this has been achieved,
further observations have no information content.
In a dynamic world, the task is to update $\bfz$ over time.
He notes that ``[the representation] need not, and probably should
not, be a detailed topographical analogue''.
MacKay also makes an interesting comparison between the tactile
domain (e.g., as experienced by a blind person using their fingers to
sense the world) and the visual domain.

One proposal for trans-saccadic fusion in the psychology literature
is the \emph{spatiotopic fusion hypothesis}, whereby information
across fixations is integrated into a high-capacity spatial buffer in
an environmental coordinate frame; see e.g., \citet{rayner-mcconkie-ehrlich-78}.
However, several experiments provide evidence against
this hypothesis, as discussed e.g., in \citet[p.\ 167]{deubel-schneider-bridgeman-02}.
For example, \citet{mcconkie-zola-79} used case alternations of word
stimuli (e.g., cHeSt and ChEsT) between parafoveal and foveal
fixations in a reading task.  They found that such changes were not
perceived and had no effect on reading performance. The phenomenon of change
blindness (see, e.g., \citealt*{rensink-oregan-clark-97})  and
the need for attention to perceive changes also provides evidence
against the spatiotopic fusion hypothesis. As an alternative,
\citet{deubel-schneider-bridgeman-02}
state that ``The current assumption is that transsaccadic
memory exists but is less image-like in form, containing more abstract
representations of the information present in each fixation.''

Despite that fact that our latent $\bfx$ representation would seem to
match the notion of a high-capacity spatial buffer, it is important to
note that it is derived from the more abstract latent representation
$\bfz$, so our model does not contradict
\citet{deubel-schneider-bridgeman-02}'s assumption stated
above.\footnote{Of course a technological solution does not have to
obey the constraints of the human visual system, but it is interesting
to make this comparison.} Instead the r\^{o}le of $\bfx$ in our model
is to enable exploitation of the geometry when relating $\bfz$ to an
observation $\bfy$. Note also that our proposed model is able to
properly handle the \emph{uncertainty} in $\bfx$ and $\bfz$ that
arises from a sequence of observations.

\paragraph{Information-theoretic criteria for saccades:}
\citet{lee-yu-00} discuss an information-theoretic framework
for understanding saccades, but their proposal deals either
with the mutual information (MI) of the activity of a hypercolumn and its
surrounding hypercolumns, or the MI of the activity of a hypercolumn
and the ``mental mosaic prediction'', where the latter is like
the spatiotopic fusion hypothesis. If we were to identify the hypercolumn with
$\bfy_j$ and the mental mosaic prediction as $p(\bfy_j|\bfz)$, then
this could be seen as an information gain criterion, but note that
the notion of the \emph{expected} IG is missing, so this would only
allow us to rank fixation locations after examining them. Also the
paper contains no implementation or experiments.

\paragraph{Visual search:} There has been a lot of work on the topic
of visual search, where the goal is for an observer to search for a
pre-defined target in the presence of a number of non-target items,
see, e.g., \citet[ch.\ 6]{findlay-gilchrist-03}. Notably 
\citet{najemnik-geisler-05} derived an ideal Bayesian observer for
this situation, where it is necessary to integrate information across
fixations, and to select where to look next. However, note that the
visual search task is very different from the one studied here;
for example, in \citet{najemnik-geisler-05} it is assumed that a
single fixation on the target is adequate to identify it, and that
the objective function driving selection of the next viewpoint
is to maximize the posterior probability of correctly identifying
the location of the target, in contrast to our EIG criterion. 

\section{Discussion \label{sec:discuss}}
In this paper we have shown how formulate the problem of fusing
multiple fixations in terms of
a high-resolution latent image $\bfx$ and linear retinal
transformations of it that yield the observed glimpses.
Combined with FA and MoFA models, this allows exact inference, and
prediction of $\bfx$. One can also learn the FA and MoFA models
from $Y$ data. We have formulated a Bayesian experimental
design problem for ``where to look next'', and given
exact results for the FA model, and bounds for the MoFA model.
We have demonstrated the models' efficacy on the
Frey faces and MNIST datasets.

There are a number of ways in which this work could be extended.
Firstly, one could consider a deep generative model (DGM) for
$p_{\theta}(\bfx|\bfz)$, where $\theta$ denotes the parameters of the
model. This will readily accept the linear adapter $V_{\ell(a)}$ to make it a
model for $p(\bfy_a|\bfz)$. However, inference for $p(\bfz|\bfy_a)$ is
more difficult in this case. A standard approach with variational
autoencoders is to have an encoder model $q(\bfz|\bfx)$ which
approximates the true posterior $p(\bfz|\bfx)$
\citep{kingma-welling-14}. However, when retinal transformations are
present, one would need an encoder for each location $a$, or (better)
one encoder that takes $\bfy_a$ and $\ell(a)$ as input. One would
then need to use $q(\bfz|\bfy_a, \ell(a))$  to approximate the EIG;
for example \citet{rainforth-foster-ivanova-bickford-smith-24} show
that the EIG can be upper bounded using a nested Monte Carlo estimator
(their eq.\ 8).

Secondly, the retinal transformation model used here allows 2D shifts of
fixation, corresponding to fronto-parallel transformations. Above we
have discussed work (e.g., \citealt*{eslami-gqn-18};
\citealt*{jang-agapito-21}) that allows more general geometric
transformations, specified by 3D locations and viewing directions. It
would be interesting to try to include variable-resolution sensors and
Bayesian experimental design into such models.

Thirdly, the data used in the current experiments consist of a single object
(face or digit). It would be very interesting to extend the work to
cover richer scenes with multiple objects; there is human experimental
data on the sequence of fixations in such images. This would require
not only latent-variable models of multiple objects, but of their
co-occurrences and inter-relationships. See \citet{williams-24} for a
discussion of structured generative models, which are one way to
approach this modelling task, and ATISS \citep{paschalidou-atiss-21}
and SceneHGN \citep{gao-sun-mo-lai-guibas-yang-23} for examples of
specific scene models.

\subsection*{Acknowledgments}
I thank Michael Gutmann, Neil Lawrence, Klaus Greff, Mike Mozer and Sjoerd van
Steenkiste for helpful discussions, and Oisin Mac Aodha and Bob Fisher
for comments on the manuscript. I thank the anonymous reviewers for
their comments that helped improve the manuscript.

For the purpose of open access, the author has applied a Creative
Commons Attribution (CC BY) licence to any Author Accepted Manuscript
version arising from this submission.

\appendix

\section{Appendix: EIG for Mixtures of Factor Analyzers \label{sec:oed-mofa}}
From eq.\ \ref{eq:eig3} we have that 
$\mathrm{EIG} = \bbE_{p(\bfz) p(\bfy_{\xi}| \bfz)} [\log
  p(\bfy_{\xi}|\bfz) - \log p(\bfy_{\xi}) ]$.
For the mixture model, the latent variables are the multivariate Gaussian
variables $\bfz^1, \ldots, \bfz^m$ (one for each factor analyzer), and
a discrete variable $c$ (mnemonic for component). Hence the first
term is
\begin{equation}
I_1  = \sum_{m=1}^M p(c=m)  \int \prod_{k=1}^M p(\bfz^k) \int
p(\bfy_{\xi}|\bfz^{1:M}, c=m)
\log p(\bfy_{\xi}|\bfz^{1:M}, c=m)) d\bfy_{\xi} \;
\prod_{k=1}^M d\bfz^k .
\end{equation}
This can be simplified by noting that $p(\bfy_{\xi}|\bfz^{1:M},
c=m) = p(\bfy_{\xi}|\bfz^m, c=m)$, i.e., that in the $m$th component,
only $\bfz^m$ is relevant, to give
\begin{equation}
I_1  = \sum_{m=1}^M \pi_m \int p(\bfz^m) \int p(\bfy_{\xi}|\bfz^m, c=m)
\log p(\bfy_{\xi}|\bfz^m, c=m) d\bfy_{\xi} \; d\bfz^m .
\end{equation}
The inner integration is just the negative entropy of $\bfy_{\xi}$
given $\bfz^m$, which arises from the noise term with covariance $\Psi^{y,m}_{\xi}$.
This Gaussian has entropy $\frac{D_{\xi}}{2} \log(2 \pi e) + \frac{1}{2}  \log |\Psi^{y,m}_{\xi}|$,
where $D_{\xi}$ is the dimensionality of $\bfy_{\xi}$. 
As this entropy does not depend on the value of $\bfz^m$, we have that
\begin{equation}
I_1  = - \sum_{m=1}^M \pi_m \left[ \frac{D_\xi}{2} \log(2 \pi e) + \frac{1}{2} \log
  |\Psi^{y,m}_{\xi}| \right] .
\end{equation}

The second term in the expression for the EIG is the entropy of the
mixture $p(\bfy_{\xi})$. In general this is analytically intractable,
but an upper bound is given in Theorem 3 of
\citet{huber-bailey-durrant-whyte-hanebeck-08}, i.e.\
\begin{equation}
 H[(\bfyxi)] \le  \sum_{m=1}^M \pi_m [ - \log \pi_m + \frac{D_{\xi}}{2}
   \log(2 \pi e) + \frac{1}{2} \log | W^m_{\xi} (W^m_{\xi} )^T +
   \Psi^{y,m}_{\xi}| \; ].
\end{equation}
   Putting the expressions for $I_1$ and $H[(\bfyxi)]$ together, we obtain
\begin{equation} 
\mathrm{EIG(mix)} \le  H(\pi) + \frac{1}{2} \sum_{m=1}^M \pi_m [\log | W^m_{\xi} (W^m_{\xi} )^T +
\Psi^{y,m}_{\xi}| - \log |\Psi^{y,m}_{\xi}| \; ],
\end{equation}
where $H(\pi) = - \sum_{m=1}^M \pi_m \log \pi_m$.  This bound is tight
when the Gaussians are well separated. Using eq.\ \ref{eig:y} for the
$\mathrm{EIG}(\bfz|\xi)$ for a single Gaussian, we have that
\begin{equation} 
\mathrm{EIG(mix)} \le  H(\pi) + \sum_{m=1}^M \pi_m \mathrm{EIG}_m(\bfz^m|\xi) .
\end{equation}

As with the single Gaussian model in sec.\ \ref{sec-oed}, the above
analysis also works for $J > 1$ observations, by using the extended
vector $\tilde{\bfy}$ as in eq.\ \ref{eq:multobs} with parameters
$\{ \tW^m_{\xi} \}_{m=1}^M$ and $\{ \tilde{\Psi}^{y,m}_{\xi}
  \}_{m=1}^M$. For $J = 2$  it is quite reasonable to do the search over
all combinations, but for larger $J$ it would make sense to do this
greedily.

\section{Derivatives of the log likelihood \label{sec:dLL}}
The log likelihood $L$ is given in eq.\ \ref{eq:ll}. For convenience
we consider $J = -L$, and differentiate the two terms in
eq.\ \ref{eq:ll} in turn. We make heavy use of
\citet{petersen-petersen-12} which gives many useful matrix derivatives;
equation $n$ therein is referenced as PP$n$.

We first consider derivatives wrt $W$ of
\begin{equation}
  J_1 = \frac{1}{2} \sum_{i=1}^n (\bfy^i)^T (V_{\ell(i)} W W^T
  V^T_{\ell(i)} + \Psi^y_{\ell(i)})^{-1} \bfy^i .
\end{equation}
Let $M_i = (V_{\ell(i)} W W^T V^T_{\ell(i)} + \Psi^y_{\ell(i)})$. Then
using PP59 we obtain
\begin{equation}
  \frac{\partial J_1}{\partial W}  = \frac{1}{2} \sum_{i=1}^n 
  (\bfy^i)^T M_i^{-1} V_{\ell(i)} \frac{\partial (W W^T)}{\partial W}
  V_{\ell(i)}^T M_i^{-1}  \bfy^i .
\end{equation}
Letting $\bfu^i = V_{\ell(i)}^T M_i^{-1} \bfy^i$  we have
\begin{equation}
  \frac{\partial J_1}{\partial W}  = \frac{1}{2} \sum_{i=1}^n 
  (\bfu^i)^T \frac{\partial (W W^T)}{\partial W} \bfu^i .
\end{equation}
Using PP70 and PP71, we obtain
\begin{equation}
  \frac{\partial J_1}{\partial W}  = - \sum_{i=1}^n  (\bfu^i
  (\bfu^i)^T ) W .
\end{equation}

Now consider
\begin{equation}
 J_2 =  \frac{1}{2} \sum_{i=1}^n \log |V_{\ell(i)} W W^T V^T_{\ell(i)}
 + \Psi^y_{\ell(i)}| .
\end{equation}
Using PP43 we obtain
\begin{equation}
 \frac{\partial J_2}{\partial W}  =  \frac{1}{2}  \sum_{i=1}^n
\mathrm{Tr} [M_i^{-1} V_{\ell(i)} \frac{\partial (W W^T)}{\partial W} V_{\ell(i)}^T] .
\end{equation}
Using PP111 we obtain
\begin{equation}
 \frac{\partial J_2}{\partial W}  =  \frac{1}{2}  \sum_{i=1}^n (V_{\ell(i)}^T
 M_i^{-1} V_{\ell(i)}) W .
\end{equation}
  
Now for derivatives wrt $\Psi^y_k$. This will pick out terms in
the sum where $\ell(i) = k$. Let the $j$th diagonal
entry in this matrix be denoted $\psi_{jj}^k$. Then using PP59 and PP73
\begin{equation}
  \frac{\partial J_1}{\partial \psi_{jj}^k} = - \frac{1}{2}
  \sum_{\ell(i)=k} (\bfy^i)^T  M_i^{-1} J_{jj} M_i^{-1} \bfy^i 
\end{equation}
where $J_{jj}$ is equal to 1 in the $jj$th entry, and 0 everywhere
else. Letting $\bfs^i = M_i^{-1} \bfy^i$, we have
\begin{equation}
  \frac{\partial J_1}{\partial \psi_{jj}^k} = - \frac{1}{2}
  \sum_{\ell(i)=k} (\bfs^i)^T  J_{jj} \bfs^i  = - \frac{1}{2}
  \sum_{\ell(i)=k} ((\bfs^i)_j)^2 ,
\end{equation}
where $(\bfs^i)_j)$ denotes the $j$th entry of $\bfs^i$.

\begin{equation}
  \frac{\partial J_2}{\partial \psi_{jj}^k} = \frac{1}{2}
  \sum_{\ell(i)=k} \frac{\partial}{\partial \psi^k_{jj}}
| V_{\ell(i)} W W^T V^T_{\ell(i)} + (\Psi^y)^k | .
\end{equation}
Using PP43 we have
\begin{equation}
  \frac{\partial J_2}{\partial \psi_{jj}^k} = \frac{1}{2} \sum_{\ell(i)=k}
  \mathrm{Tr}[ M_i^{-1} J_{jj} ] = \frac{1}{2} \sum_{\ell(i)=k}
  (M_i^{-1})_{jj} .
\end{equation}
To ensure the non-negativity of $\psi^k_{jj}$ we parameterize it as
$\psi^k_{jj} = \exp(t^k_j)$, where $t^k_j$ is a real number.

\bibliographystyle{apalike}

\end{document}